\documentclass[runningheads]{llncs}

\usepackage{eccv}

\usepackage{eccvabbrv}

\usepackage{graphicx}
\usepackage{booktabs}

\usepackage{multirow}
\usepackage{color}
\usepackage{colortbl}
\usepackage{algorithm}
\usepackage{algpseudocode}

\usepackage[accsupp]{axessibility}  

\usepackage{hyperref}

\usepackage{orcidlink}

\begin{document}

\title{Learn to Preserve and Diversify: Parameter-Efficient Group with Orthogonal Regularization for Domain Generalization} 

\titlerunning{Parameter-Efficient Group with Orthogonal Regularization for DG}

\author{Jiajun Hu\inst{1,2},
Jian Zhang\inst{1,2},
Lei Qi\inst{3,*},
Yinghuan Shi\inst{1,2,*},
Yang Gao\inst{1,2}}

\renewcommand{\thefootnote}{}
\footnotetext{* Corresponding Authors: Yinghuan Shi (\email{syh@nju.edu.cn}) and Lei Qi (\email{qilei@seu.edu.cn}).}

\authorrunning{J.~Hu et al.}

\institute{State Key Laboratory for Novel Software Technology, Nanjing University, China \and
National Institute of Healthcare Data Science, Nanjing University, China \and
School of Computer Science and Engineering, Key Lab of Computer Network and Information Integration (Ministry of Education), Southeast University, China}

\maketitle

\begin{abstract}
  Domain generalization (DG) aims to avoid the performance degradation of the model when the distribution shift between the limited training data and unseen test data occurs. Recently, foundation models with enormous parameters have been pre-trained with huge datasets, demonstrating strong generalization ability and showing promising direction for solving the DG problem. However, fully Fine-Tuning (FT) the foundation models results in unsatisfactory out-of-distribution accuracy due to the destroyed pre-trained generalized features. Recently, Parameter-Efficient Fine-Tuning (PEFT) alleviates the above problem by fine-tuning a small portion of the model parameters while keeping the rest frozen, which achieves better generalization performance compared to FT. Nevertheless, PEFT still suffers from the issue of overfitting to the training domains. To address the above issue, we propose Parameter-Efficient Group with Orthogonal regularization (PEGO) for vision transformers, which effectively preserves the generalization ability of the pre-trained network and learns more diverse knowledge compared with conventional PEFT. Specifically, we inject a group of trainable Low-Rank Adaptation (LoRA) modules into the pre-trained model and propose an orthogonal regularization loss to enhance the generalization ability of the model. Our framework achieves SOTA performance on five DG benchmarks, while only requiring training a small number of parameters without adding additional testing cost.
\end{abstract}

\section{Introduction}
Traditional machine learning algorithms assume that training data and test data come from independent and identical distributions~\cite{iid}. 
However, a trained model suffers significant performance degradation when the distribution discrepancy (\textit{a.k.a} domain gap) between the training and test data is large.
To address this issue, Domain Generalization (DG) is proposed, which assumes that the model trained with only source domains can generalize well in the unseen target domains. Previous DG works~\cite{irm,IBIRM,Domaindrop,dann,cdann} mostly aim to facilitate model generalizability by either extracting domain-invariant features that are applicable across domains or augmenting training source domains with manually defined transformations, \eg, style transfer~\cite{mixstyle,wang2021learning}, Fourier transformation~\cite{fact,ALOFT}, \etc.

However, most DG works follow the same training strategy that directly fine-tunes a pre-trained model  (\eg, ResNet~\cite{resnet} pre-trained on ImageNet~\cite{imagenet}) without considering the influence of the initial parameters on the ultimate generalization performance of finally-trained models.  
For example, we find that when using a randomly initialized ResNet-50 as the backbone, the average performance of Empirical Risk Minimization (ERM that trains a model by simply aggregating all source domains without any other techniques) on the PACS~\cite{pacs} dataset is only 35.8\%. This result is much lower than the performance of ResNet-50 pre-trained on ImageNet, which is 84.2\%. Particularly, on the domain \textit{sketch} in PACS, where the distribution is far from ImageNet, fine-tuning the pre-trained ResNet-50 has a 55.3\% performance improvement (24.0\% $\rightarrow$ 79.3\%) than fine-tuning the randomly initialized ResNet-50. This indicates that \textit{the generalizable knowledge from pre-trained models should be fully exploited for better generalization for downstream DG tasks.}

Recently, due to the flourishing development of deep learning, both the parameters and training data of deep models have largely increased and everyone can easily access these pre-trained large foundation models~\cite{foundation} based on the vision transformer architecture (ViT)~\cite{vit}.
Previous works~\cite{broad_dg,zhang2022delving,sparse} have shown that the vision transformer is more robust to unknown distributions than CNN~\cite{CNN}, and CLIP~\cite{clip} model trained on 400M image-text pairs has demonstrated strong zero-shot generalization ability. 
However, a critical question arises that \textit{does directly fine-tuning stronger models lead to better results on DG tasks?} The answer is \textbf{NO}. Cha \etal~\cite{miro} found that on the DomainBed~\cite{domainbed} benchmark, the performance of fine-tuning ViT-B~\cite{vit} pre-trained from CLIP~\cite{clip} is lower than that of fine-tuning ResNet-50 pre-trained on ImageNet (ViT-B: 61.1\% \vs ResNet-50: 64.2\%).
This is because direct fine-tuning distorts the generalizable features that originally reside in the pre-trained model~\cite{LPFT}. 
Compared to the small models,
a large number of parameters of the foundation models cause more severe overfitting issues when training with limited source domain data. 
Previous DG methods~\cite{irm,IBIRM,Domaindrop,dann,cdann} mainly focus on how to extract domain-invariant features from limited source domains or perform data augmentation to generate more training data, ignoring how to preserve and exploit the generalization ability of the pre-trained models itself to improve the out-of-distribution generalization performance.
Furthermore, with a huge number of parameters in foundation models, fine-tuning these parameters requires high training overhead in both GPU memory and training time, which significantly increases the difficulty of successfully fine-tuning a foundation model for users with limited resources.

To address the abovementioned two issues, Parameter-Efficient Fine-Tuning (PEFT)~\cite{adapter,lora,vpt} has attracted significant interest in various language and visual tasks. Compared to full Fine-Tuning (FT), PEFT methods inject lightweight trainable modules into the pre-trained model and freeze all the parameters of the pre-trained model. This approach reduces training overhead and achieves comparable or better performance than FT on downstream tasks~\cite{adapter}. 
Low-Rank Adaptation (LoRA)~\cite{lora} is one of the most commonly employed PEFT implementations, which injects trainable rank decomposition matrices into every layer of the transformer~\cite{transformer}.
Moreover, we discover that LoRA demonstrates substantial performance enhancement in addressing out-of-distribution tasks compared to FT and LoRA outperforms some conventional DG algorithms 
(In \cref{sec:main_results}).

Despite the advantages of little computational overhead and overfitting alleviation, applying LoRA for DG bears two limitations. 
Firstly, although LoRA only injects little parameters into a foundation model that aims to alleviate the feature distortion problem, it still suffers from this problem since the features learned from the LoRA module may conflict with the feature of the pre-trained model, resulting in knowledge forgetting.
Secondly, LoRA only employs a single low-rank module in each layer, which also easily overfits the training domain and cannot handle various unseen domains, further limiting its generalization performance.

To address the above limitations of LoRA, we propose a novel \textbf{P}arameter \textbf{E}fficient \textbf{G}roup with \textbf{O}rthogonal regularization (\textbf{PEGO}) framework to fully exploit the potential of the pre-trained foundation models to solve the DG problem. 
First, we \textbf{Preserve} the generalization ability of the pre-trained model learned from the large-scale pre-training by imposing an orthogonal regularization loss between the pre-trained weights and the weights of LoRA layers. In this way, we can effectively minimize the distortion of the pre-trained generalized features. 
Second, we employ a group of LoRA modules for each layer to \textbf{Diversify} feature representations during training. With the learned abundant features, the model can better handle various unseen domains during the test. To further encourage diversity, the orthogonal constraints are also added between the weights of these LoRA modules.
We summarize the contributions of this work as follows:

\begin{enumerate}
    \item We propose a novel PEFT framework named PEGO that can effectively alleviate the overfitting issue with little computational overhead.
    \item We design an orthogonal regularization loss to facilitate knowledge preservation of the pre-trained model.
    \item We design the LoRA group with diversity constraints to learn diverse features to handle various unseen domains.
    \item On five DomianBed benchmarks, PEGO achieves state-of-the-art performance compared to previous DG algorithms. Moreover, our method outperforms other PFET methods and the methods exploiting pre-trained models. 
\end{enumerate}

\section{Related Work}

\subsection{Domain Generalization}
Domain generalization (DG) aims to ensure that a model trained on source domains can generalize well to any unseen target domains. There are various methods for DG problem, mainly including data augmentation ~\cite{L2A,ddaig,fact,Normaug,mixstyle,wang2022feature}, meta-learning~\cite{mldg,masf,metareg,MVDG,MEDIC}, ensemble learning~\cite{dael,swad,SMA,DiWA}, self-supervised learning~\cite{jigen,EISNet}, adversarial learning~\cite{dann,MMD,cdann}, causal learning~\cite{irm,IBIRM,matchdg}, test time adaptation~\cite{T3A,improved_TTA,zhang2022generalizable,Domainadaptor}, \etc.

Most of the previous DG works choose small models (\eg, AlexNet~\cite{AlexNet} and ResNet~\cite{resnet}) as the pre-trained backbone. Different from them, several methods exploit large pre-trained models for out-of-distribution generalization.  
MIRO~\cite{miro} proposes mutual information regularization by assuming the pre-trained model as the oracle model.
GESTUR~\cite{GESTUR} designs a gradient estimation to reduce potential risks in unseen domains utilizing a large pre-trained model. These two methods require significant training costs due to optimizing all the parameters of the pre-trained model. 
Moreover, there are some recent works~\cite{Promptstyler,car-ft,CLIPood,A_Sentence} that utilize the text information from vision-language models to enhance the generalization ability of the fine-tuned model, but these methods rely on the jointly trained text encoder and visual encoder.

\subsection{Parameter-Efficient Fine-Tuning}
Parameter-Efficient Fine-Tuning (PEFT)~\cite{adapter} is first proposed to fine-tune large pre-trained transformers in natural language processing tasks, effectively reducing the computational and storage cost. PEFT methods inject some lightweight modules into the foundation model and only optimize a small portion 
of the model parameters to achieve a similar or higher performance compared with FT on downstream tasks.
PEFT methods are also applied to deal with visual tasks. For example, VPT~\cite{vpt} introduces trainable prompt tokens in the input space for ViT~\cite{vit}. One of the most influential PEFT methods is LoRA~\cite{lora}, which inserts trainable low-rank decomposition matrices into the transformer block while freezing the pre-trained model weights. It does not introduce additional inference latency and works with any neural network with dense layers.

\subsection{Orthogonal Regularization}
Orthogonality in neural networks~\cite{xie2017all,brock2016neural,BigGAN,bansal2018can} has been widely studied to improve training stability, training speed, and the performance of the model. For example, Xie \etal~\cite{xie2017all} utilize the orthonormality among different filter banks to mitigate the gradient vanishing on training deep convolutional neural networks (CNN).
Bansal \etal~\cite{bansal2018can} develop novel orthogonality regularizations, which achieve better accuracy, and faster and more stable convergences on several CNNs. In the DG literature, there are some works~\cite{DSN,Decaug,steam,POEM} applying orthogonal loss to disentangle the domain-invariant representations and domain-specific representations, but these methods usually require designing the additional network to generate two decoupled representations. Orthogonality is also extensively used in continual learning to prevent catastrophic forgetting of past tasks~\cite{OWM,OGD,GPM,Olora}. The most relevant to our work is O-LoRA~\cite{Olora}, which utilizes orthogonal subspace learning for continual learning in language models. Different from their task, this paper aims to solve the DG problem when fine-tuning the foundation model and we additionally consider the relationship between the pre-trained weights and the injected LoRA layers.

\section{Methods}

\subsection{Problem Formulation}
For the classification task in the DG setting, the training data $D_{tr}$ usually contains $n_{d}$ domains $\{D_{i}\}_{i=1}^{n_{d}}$ ($n_{d}>1$). The test data $D_{te}$ is not accessible during training but shares the same category space as $D_{tr}$. We define that $x_{i}$ is the $i$-th sample in $D_{tr}$ with the category label $y_{i}$. The distribution of training data $P_{tr}$ is the joint distribution over image space and label space in $D_{tr}$, and there is a distribution gap between $P_{tr}$ and the distribution of test data $P_{te}$.

The entire model includes a feature extractor $F(\cdot;\theta)$ parameterized by $\theta$ and a classifier $H(\cdot;\psi)$ parameterized by $\psi$. Previous DG works usually choose a pre-trained model and fine-tune it on the distribution of training data $P_{tr}$. The goal of DG is that $F$ and $H$ trained on $D_{tr}$ can generalize well on $D_{te}$.

\subsection{Revisit Previous Methods}

\subsubsection{Fine-Tuning.}
Fine-Tuning (FT, updating all the parameters of the model) is the most common training manner in previous DG works. The optimization objective of FT can be defined as the following formula:
\begin{equation}
    \min_{\theta,\psi} \quad \mathcal{L}_{cls} = \mathcal{L}_{CE}(H(F(x;\theta);\psi), y),
\end{equation}
where $(x,y)\sim P_{tr}$ and $\mathcal{L}_{CE}$ is the cross-entropy loss. Gulrajani \etal~\cite{domainbed} claim that ERM based on FT has competitive performance compared to most DG algorithms. However, Kumar \etal~\cite{LPFT} argue that FT distorts pre-trained features, which leads to poor out-of-distribution performance. In \cref{sec:main_results}, we also find that the performances of ERM are much lower than those of other methods when choosing a foundation model as the backbone.

\subsubsection{Parameter-Efficient Fine-Tuning.}
Compared with FT, Parameter-Efficient Fine-Tuning (PEFT) only updates a small number of parameters.
For any PEFT method, the network with PEFT modules can be defined as $G(\cdot;\Phi)$ parameterized by $\Phi$, where the trainable parameters are $\phi \subseteq \Phi$ (the parameter number of $\phi$ is much smaller than that of $\Phi$). The general optimization objective of any PEFT method can be formulated as follows:
\begin{equation}
\label{eq:peft}
    \min_{\phi, \psi} \quad \mathcal{L}_{cls} = \mathcal{L}_{CE}(H(G(x; \Phi);\psi), y).
\end{equation}
Due to not updating any parameters of the foundation model, PEFT effectively inherits the generalization ability of the foundation model. However, in the DG setting, the trainable parameters in PEFT still suffer from overfitting to the source domains, resulting in a performance decrease on the unseen test domain. 
At the same time, the injected PEFT modules also partially distort the generalized features produced by the foundation model during training.

\begin{figure*}[!tbp]
  \centering
  \includegraphics[width=0.75\linewidth]{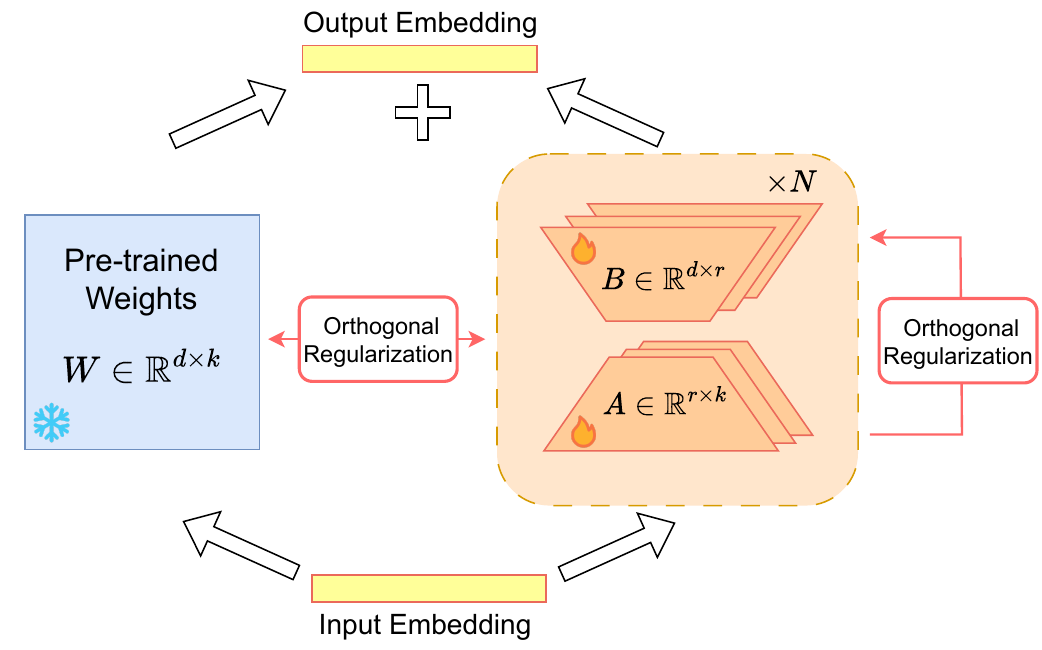}
  \caption{Illustration of our method: \textbf{P}arameter \textbf{E}fficient \textbf{G}roup with \textbf{O}rthogonal regularization (PEGO). Different from previous DG work updating all the parameters of the pre-trained model, we freeze the parameters of the model and inject a group of trainable parameter-efficient modules into it. Moreover, we apply an orthogonal regularization loss between the pre-trained weights and the LoRA modules to preserve the generalization ability of the pre-trained model (\textit{Learn to Preserve}) and employ another orthogonal regularization loss on different LoRA modules within the group to encourage them to learn diverse knowledge during training (\textit{Learn to Diversify}).}
  \label{fig:method}
\end{figure*}

\subsection{Parameter-Efficient Group with Orthogonal Regularization}
To address the issue of FT and PEFT in DG, we propose \textbf{P}arameter \textbf{E}fficient \textbf{G}roup with \textbf{O}rthogonal regularization (PEGO) based on LoRA~\cite{lora}, which is a classic PEFT method for pre-trained transformer. In this section, we first review the key technologies of LoRA and then introduce the details of our method.\\

\noindent\textbf{LoRA.}
Hu \etal~\cite{lora} propose \textbf{Lo}w-\textbf{R}ank \textbf{A}daptation (LoRA), utilizing the hypothesis that the pre-trained weights have a low intrinsic rank. Specifically, for a linear layer in the pre-trained model, which is parameterized by $W \in \mathbb{R}^{d\times k}$, there is a low-rank decomposition $W+\Delta W=W+BA$, where $B \in \mathbb{R}^{d\times r}$, $A \in \mathbb{R}^{r\times k}$ and $r \ll \min{(d, k)}$. During training, LoRA only updates injected low-rank weights $A$ and $B$ while keeping $W$ frozen. We define the input feature of the linear layer as $z_{in}$, and the output feature of the linear layer can be calculated through the following forward process:
\begin{equation}
\label{eq:forward}
    z_{out} = (W + \Delta W)z_{in} = (W + BA)z_{in} = Wz_{in} + BAz_{in}.
\end{equation}
In the original paper, the LoRA layers are only injected into the $W^{q}$ and $W^{v}$, which are the query and value projection matrices of the self-attention modules. After training, we can directly compute $W_{final} = W + BA$ as the final weight. Therefore, there is no additional inference latency in LoRA.\\

\noindent\textbf{PEGO.}
To further improve the out-of-distribution performance of LoRA, we propose the idea of learning to \textbf{Preserve} and \textbf{Diversify}. The former aims to preserve the generalization performance of the pre-trained model during fine-tuning, while the latter aims to learn more diverse knowledge from source domains. Specifically, we propose to inject a group of LoRA modules into the pre-trained network and apply an orthogonal regularization loss to achieve the above two goals simultaneously. \cref{fig:method} provides an illustration of our method.
\newline
\textbf{\textbullet~ Learn to Preserve}

Inspired by the orthogonal gradient updating~\cite{OWM,OGD,GPM} used to prevent catastrophic forgetting, we strive to constrain the gradient subspace of fine-tuning to be orthogonal to the gradient subspace of large-scale pre-training task. This enables the model to learn useful information in the source domains while preserving the generalization ability of the pre-trained model. However, we may not be able to access the pre-training dataset and it is not practical to calculate the full gradient of the large-scale pre-training task. Following Wang \etal~\cite{Olora} who regard the weights of the LoRA layer as the gradient subspace of a certain task, we propose to consider the original pre-trained weights as the gradient subspace of the pre-training task. Similarly, the fine-tuned LoRA weights can be considered as the gradient subspace updated on the source domains. With these two gradient subspaces,
we propose an orthogonal regularization loss to constrain the pre-trained weights $W$ orthogonal to the weights of injected module $BA$. Specifically, the loss can be formulated as follows:
\begin{equation}
    \mathcal{L}_{preserve} = \left\Vert W^{T}(BA) \right\Vert_{1},
\end{equation}
where $\Vert\cdot\Vert_{1}$ is $L1$ norm. As the same as LoRA, we only fine-tune the low-rank weight matrices $A$ and $B$ while keeping the rest parameters frozen during training. 

Furthermore, we analyze the above loss from the perspective of the feature-level. According to the forward process in \cref{eq:forward}, we define the output feature of the pre-trained weight $z_{init}$ = $Wz_{in}$ and the output feature of the LoRA layer $z_{new}$ = $BAz_{in}$. While the loss restricts $BA$ orthogonal to $W$, it indirectly constrains $z_{init}$ orthogonal to $z_{new}$. This can be demonstrated by the following transpose transformation:
\begin{equation}
    z_{init}^{T}z_{new} = (Wz_{in})^{T} BAz_{in} = z_{in}^{T} (W^{T}BA) z_{in}.
\end{equation}
Since the features generated by fine-tuning are encouraged to be orthogonal to pre-trained features, the generalization ability of the pre-trained model is preserved well. At the same time, during the implementation, we discovered that optimizing weights to be orthogonal requires fewer computational resources and results in better performance compared to optimizing features to be orthogonal.
\newline
\textbf{\textbullet~ Learn to Diversify}

In the DG literature, increasing the diversity of the training trajectories is used to improve the generalization performance of the model. For example, Zhang \etal~\cite{MVDG} propose a multi-view algorithm for employing multiple optimization trajectories; Arpit \etal~\cite{SMA} perform the ensemble of multiple independently trained models. However, these methods require additional multi-step gradient updates or training multiple models. Benefiting from the lightweight and easily scalable characteristics of LoRA, we propose to introduce multiple LoRA modules and apply orthogonal regularization to facilitate the model learning diverse knowledge.


Different from the original LoRA where the pre-trained weight $W$ comes with only one trainable low-rank matrices $BA$, we employ a parameter-efficient group of LoRA layers $g = \{A_{i},B_{i}\}_{i=1}^{N}$ in our framework, where $N$ is the number of LoRA layers. Moreover, we adopt a pairwise orthogonal regularization loss to enhance the diversity of knowledge learned by each LoRA layer. Specifically, for a certain LoRA layer $g_{i} = \{A_{i},B_{i}\}$, its weight matrix $B_{i}A_{i}$ is encouraged to be orthogonal to the weight matrix of other LoRA layers $\{B_{j}A_{j}\}_{j\ne i}^{N}$. Formally, the loss that aims to learn to diversify is defined as follows:
\begin{equation}
    \mathcal{L}_{diversify} = \sum_{i=1}^{N}\sum_{j=i+1}^{N}\left\Vert (B_{i}A_{i})^{T}(B_{j}A_{j}) \right\Vert_{1}.
\end{equation}
The above loss promotes the orthogonality between the weights of different LoRA modules, resulting in the output features of different LoRA modules are also encouraged to orthogonal. Our method learns more diverse optimization trajectories compared to the original LoRA and only increases little training cost.
\newline
\textbf{\textbullet~ Final Object}

We combine $\mathcal{L}_{preserve}$ and $\mathcal{L}_{diversify}$ as the optimization objective of orthogonal regularization, which takes the following form:
\begin{equation}
\label{eq:L_O}
    \mathcal{L}_{O}(W) = \sum_{i=1}^{N}\left\Vert W^{T}(B_{i}A_{i}) \right \Vert_{1} + \sum_{i=1}^{N}\sum_{j=i+1}^{N}\left\Vert (B_{i}A_{i})^{T}(B_{j}A_{j}) \right\Vert_{1}.
\end{equation}
In line with LoRA, we only apply the above loss to $W^{q}$ and $W^{v}$ in each block of the pre-trained transformer. During training, we only fine-tune the LoRA group and the classification head while keeping the rest parameters frozen. The final orthogonal regularization loss can be given by:
\begin{equation}
   \mathcal{L}_{OR} = \sum_{b=1}^{B}(\mathcal{L}_{O}(W^{q}_{b}) + \mathcal{L}_{O}(W^{v}_{b})),
\end{equation}
where $B$ is the number of blocks in ViT, $W^{q}_{b}$ and $W^{v}_{b}$ indicates the query and value projection matrices in the $b$-th block. Finally, we combine $\mathcal{L}_{cls}$ with $\mathcal{L}_{OR}$ as the final optimization object of the model:
\begin{equation}
\label{eq:final}
   \mathcal{L}_{final} = \mathcal{L}_{cls} + \alpha\mathcal{L}_{OR},
\end{equation}
where $\alpha$ is the balancing coefficient of two losses. When the model is deployed to the test environment after training, we merge the group of LoRA layers to the pre-trained weight:
\begin{equation}
    W_{final} = W + \sum_{i=1}^{N}B_{i}A_{i}.
\label{eq:merge}
\end{equation}
As the same as LoRA, there is no additional testing latency in our method. 

\section{Experiments}

\subsection{DataSets}
We use five common datasets in DomainBed~\cite{domainbed} evaluation benchmarks to verify the effectiveness of our method:
\textbf{1. PACS}~\cite{pacs} includes 9,991 images, 7 categories, and 4 domains. The domain shift between each domain is large (\eg, \textit{photo} and \textit{sketch}).
\textbf{2. VLCS}~\cite{vlcs} includes 10,729 images, 5 categories, and 4 domains. The domain shift between each domain is mainly from different viewpoints.
\textbf{3. OfficeHome}~\cite{officehome} includes 15,500 images, 65 categories, and 4 domains. It contains more categories and smaller domain shift than PACS.
\textbf{4. TerraIncognita}~\cite{terra} includes 24,788 images, 10 categories, and 4 domains. The images are taken in four different wild locations and it's a challenging dataset.
\textbf{5. DomainNet}~\cite{domainnet} includes 586,575 images, 345 categories, and 6 domains. Its number of images and categories far exceeds the above datasets.

\subsection{Implementation Details}
Different from DomainBed~\cite{domainbed} using ResNet-50~\cite{resnet} as the backbone, we utilize ViT-B/16~\cite{vit} pre-trained by CLIP~\cite{clip} as the default foundation model. During training, we employ the Adam~\cite{Adam} optimizer with a learning rate of 5e-4 for 5000 iterations, except for the DomainNet, which requires 15000 iterations for convergence. The 32 images in each source domain construct a whole batch. Our data augmentation includes random horizontal flip, color jittering, and random graying in DomainBed. We set the rank of LoRA $r$ to 4 and the balancing coefficient $\alpha$ to 1e-3. For the numbers of LoRA layers $N$ in each group, we define the hyperparameter search space as $N \in \{2, 4, 6\}$. More details about the evaluation protocol and hyperparameters search can be found in Supplementary.

\subsection{Main Results}
\label{sec:main_results}

\begin{table*}[!tbp]
  \centering
  \caption{Performance comparison with DG methods. Leave-one-domain-out accuracy (\%) on five DomainBed benchmarks. In addition to the results of our method, other results come from Lew \etal~\cite{GESTUR}. OH, TI and DN indicate OfficeHome, TerraIncognita, and DomainNet, respectively (similarly hereinafter).}
  \scalebox{0.93}{
  \begin{tabular}{p{2.2cm}<{\centering}|p{1.5cm}<{\centering}p{1.5cm}<{\centering}p{1.5cm}<{\centering}p{1.5cm}<{\centering}p{1.5cm}<{\centering}|p{0.8cm}<{\centering}}
    \toprule
    Algorithm & PACS & VLCS & OH & TI & DN & Avg\\
    \midrule
    ERM (FT) & 83.4{\scriptsize$\pm$0.5} & 75.9{\scriptsize$\pm$1.3} & 66.4{\scriptsize$\pm$0.5} & 35.3{\scriptsize$\pm$0.8} & 44.4{\scriptsize$\pm$0.6} & 61.1 \\
    SWAD~\cite{swad} & 91.3{\scriptsize$\pm$0.1} & 79.4{\scriptsize$\pm$0.4} & 76.9{\scriptsize$\pm$0.1} & 45.4{\scriptsize$\pm$0.5} & 51.7{\scriptsize$\pm$0.8} & 68.9 \\
    SMA~\cite{SMA} & 92.1{\scriptsize$\pm$0.2} & 79.7{\scriptsize$\pm$0.2} & 78.1{\scriptsize$\pm$0.1} & 48.3{\scriptsize$\pm$0.7} & 55.9{\scriptsize$\pm$0.2} & 70.8 \\
    MIRO~\cite{miro} & 95.6{\scriptsize$\pm$0.8} & 82.2{\scriptsize$\pm$0.3} & 82.5{\scriptsize$\pm$0.1} & 54.3{\scriptsize$\pm$0.4} & 54.0{\scriptsize$\pm$0.3} & 73.7\\
    GESTUR~\cite{GESTUR} & 96.0{\scriptsize$\pm$0.0} & 82.8{\scriptsize$\pm$0.1} & \textbf{84.2{\scriptsize$\pm$0.1}} & 55.7{\scriptsize$\pm$0.2} & 58.9{\scriptsize$\pm$0.1} & 75.5 \\
    \midrule
    \rowcolor{blue!10} Ours & \textbf{96.5{\scriptsize$\pm$0.1}} & \textbf{83.2{\scriptsize$\pm$0.3}} & \textbf{84.2{\scriptsize$\pm$0.1}} & \textbf{57.3{\scriptsize$\pm$0.3}} & \textbf{59.3{\scriptsize$\pm$0.1}} &
    \textbf{76.1} \\
    \bottomrule
  \end{tabular}
  }
  \label{tab:DG}
\end{table*}

\textbf{Comparison with DG Methods.}
We first compare PEGO with the baseline ERM (\ie, FT) and state-of-the-art (SOTA) DG methods, including: SWAD~\cite{swad}, SMA~\cite{SMA}, MIRO~\cite{miro} and GESTUR~\cite{GESTUR}. SWAD and SMA are both ensemble methods that show significant improvement compared to ERM on DomainBed using ResNet-50 as the backbone; MIRO and GESTUR both aim to preserve and exploit the generalization ability of the pre-trained network.

The results of the performance comparison are shown in \cref{tab:DG}.
Compared to ERM, all DG approaches have significantly improved, indicating much potential for further enhancement when using foundation models for the DG problem. Furthermore, PEGO outperforms all previous methods on all five datasets, demonstrating the superiority of our method. Especially on the challenging TerraIncognita dataset, PEGO achieves a remarkable improvement which is more than 1.6\% compared with other methods (55.7\% $\rightarrow$ 57.3\%).

\begin{table*}[!tbp]
  \centering
  \caption{Performance comparison with PEFT methods. Leave-one-domain-out accuracy (\%) on five DomainBed benchmarks.}
  \scalebox{0.93}{
  \begin{tabular}{p{2.2cm}<{\centering}|p{1.5cm}<{\centering}p{1.5cm}<{\centering}p{1.5cm}<{\centering}p{1.5cm}<{\centering}p{1.5cm}<{\centering}|p{0.8cm}<{\centering}}
    \toprule
    Algorithm & PACS & VLCS & OH & TI & DN & Avg\\
    \midrule
    ERM (FT) & 83.4{\scriptsize$\pm$0.5} & 75.9{\scriptsize$\pm$1.3} & 66.4{\scriptsize$\pm$0.5} & 35.3{\scriptsize$\pm$0.8} & 44.4{\scriptsize$\pm$0.6} & 61.1 \\
    Adapter~\cite{adapter} & 92.0{\scriptsize$\pm$0.5} & 79.8{\scriptsize$\pm$0.4} & 72.9{\scriptsize$\pm$0.4} & 44.4{\scriptsize$\pm$0.8} & 56.2{\scriptsize$\pm$0.1} & 69.1 \\
    LoRA~\cite{lora} & 96.0{\scriptsize$\pm$0.1} & 82.7{\scriptsize$\pm$0.0} & 83.4{\scriptsize$\pm$0.1} & 54.8{\scriptsize$\pm$0.6} & 58.1{\scriptsize$\pm$0.1} & 75.0 \\
    VPT~\cite{vpt} & 96.2{\scriptsize$\pm$0.3} &  82.9{\scriptsize$\pm$0.3} & 83.4{\scriptsize$\pm$0.3} & 54.2{\scriptsize$\pm$0.7} & 58.9{\scriptsize$\pm$0.1} & 75.1 \\
    \midrule
    \rowcolor{blue!10} Ours & \textbf{96.5{\scriptsize$\pm$0.1}} & \textbf{83.2{\scriptsize$\pm$0.3}} & \textbf{84.2{\scriptsize$\pm$0.1}} & \textbf{57.3{\scriptsize$\pm$0.3}} & \textbf{59.3{\scriptsize$\pm$0.1}} &
    \textbf{76.1} \\
    \bottomrule
  \end{tabular}
  }
  \label{tab:PEFT}
\end{table*}

\noindent\textbf{Comparison with PEFT Methods.}
In this subsection, we make a performance comparison between PEGO and several PEFT methods, containing Adapter~\cite{adapter} and LoRA~\cite{lora} which are widely utilized in language tasks, VPT~\cite{vpt} which is designed specifically for vision transformer.

As shown in \cref{tab:PEFT}, the performances of all PEFT methods on five DomainBed benchmarks are significantly higher than that of FT. This indicates that full fine-tuning considerably harms the generalization performance of the pre-trained model, while PEFT can effectively address this issue. Moreover, our method achieves state-of-the-art performance on five datasets and 1.1\% improvement of average performances compared to LoRA which is the basis of our method (75.0\% $\rightarrow$ 76.1\%).

\begin{table*}[!tbp]
  \centering
  \caption{Performance comparison with methods exploiting pre-trained models. Leave-one-domain-out accuracy (\%) on four DomainBed benchmarks. The results of WiSE-FT come from Lew \etal~\cite{GESTUR} and we report the rest results. The best and second-best accuracy are \textbf{bolded} and \underline{underlined}, respectively.}
  \scalebox{0.93}{
  \begin{tabular}{p{2.3cm}<{\centering}|p{1.5cm}<{\centering}p{1.5cm}<{\centering}p{1.5cm}<{\centering}p{1.5cm}<{\centering}|p{0.8cm}<{\centering}}
    \toprule
    Algorithm & PACS & VLCS & OH & TI & Avg\\
    \midrule
    ERM (FT) & 83.4{\scriptsize$\pm$0.5} & 75.9{\scriptsize$\pm$1.3} & 66.4{\scriptsize$\pm$0.5} & 35.3{\scriptsize$\pm$0.8} & 65.3 \\
    L$^{2}$-SP~\cite{L2SP} & 92.2{\scriptsize$\pm$0.7} & 81.0{\scriptsize$\pm$0.2} & 68.2{\scriptsize$\pm$0.5} & 39.4{\scriptsize$\pm$1.6} & 70.2 \\
    LP-FT~\cite{LPFT} & 94.2{\scriptsize$\pm$0.7} & 77.5{\scriptsize$\pm$0.4} & 72.0{\scriptsize$\pm$0.4} & 39.0{\scriptsize$\pm$1.5} & 70.7 \\
    LwF~\cite{LwF} & 93.6{\scriptsize$\pm$0.6} & 81.9{\scriptsize$\pm$0.4} & 80.7{\scriptsize$\pm$0.4} & 39.4{\scriptsize$\pm$0.6} & 73.9 \\
    WiSE-FT~\cite{wise} & \underline{94.5{\scriptsize$\pm$0.0}} & \textbf{83.9{\scriptsize$\pm$0.3}} & \underline{83.9{\scriptsize$\pm$0.2}} & \underline{47.5{\scriptsize$\pm$1.2}} & \underline{77.5} \\
    \midrule
    \rowcolor{blue!10} Ours & \textbf{96.5{\scriptsize$\pm$0.1}} & \underline{83.2{\scriptsize$\pm$0.3}} & \textbf{84.2{\scriptsize$\pm$0.1}} & \textbf{57.3{\scriptsize$\pm$0.3}} & \textbf{80.3} \\
    \bottomrule
  \end{tabular}
  }
  \label{tab:CL}
\end{table*}

\noindent\textbf{Comparison with Methods Exploiting Pre-trained Models.}
There are some works in other fields utilizing pre-trained models to improve the generalization ability of the fine-tuned model. Following Cha \etal~\cite{miro}, we select several classic methods to make a performance comparison with PEGO. Specifically, L$^{2}$-SP~\cite{L2SP} employs $L^{2}$ penalty between the pre-trained model and fine-tuned model during training; LP-FT~\cite{LPFT} proposes a strategy of first linear probing and then full fine-tuning; LwF~\cite{LwF} constrains the outputs of fine-tuned model for old tasks to be similar to that of pre-trained network; WiSE-FT~\cite{wise} ensembles the weights of the zero-shot and fine-tuned networks to realize robust fine-tuning.

We report the results of all methods on four DomainBed benchmarks in \cref{tab:CL}.
Our method achieves the best performance on PACS, OfficeHome, and TerraIncognita, except for VLCS, where it is only 0.7\% lower than WiSE-FT. Furthermore, PEGO outperforms the previous best method by 2.8\% on the average accuracy of four benchmarks (77.5\% $\rightarrow$ 80.3\%). This is mainly because PEGO has over 9.8\% improvement on the TerraIncognita dataset (47.5\% $\rightarrow$ 57.3\%), and its standard error is significantly smaller. The above results indicate that although previous methods achieve some improvement compared with FT and aim to preserve the generalization ability of pre-trained models, there is still a significant overfitting phenomenon when using these methods. Our method effectively alleviates the above issue.

\section{Further Analysis}

\begin{table*}[!tbp]
  \centering
  \caption{Ablation study on two orthogonal regularization losses. Leave-one-domain-out accuracy (\%) on PACS and OfficeHome.}
  \scalebox{0.93}{
  \begin{tabular}{p{2cm}<{\centering}p{1.8cm}<{\centering}p{1.8cm}<{\centering}|p{2cm}<{\centering}p{2cm}<{\centering}}
    \toprule
    \multicolumn{1}{c|}{} & $\mathcal{L}_{preserve}$ & $\mathcal{L}_{diversify}$ & PACS & OH \\
    \midrule
    \multicolumn{1}{p{1.8cm}<{\centering}|}{\multirow{4}{*}{PEGO}} & \cellcolor{blue!10} \checkmark & \cellcolor{blue!10}\checkmark & \cellcolor{blue!10}\textbf{96.55{\scriptsize$\pm$0.11}} & \cellcolor{blue!10}\textbf{84.21{\scriptsize$\pm$0.10}} \\
    \multicolumn{1}{c|}{} & \checkmark & & 96.14{\scriptsize$\pm$0.25} & 82.95{\scriptsize$\pm$0.02} \\
    \multicolumn{1}{c|}{} & \multicolumn{1}{c}{} & \checkmark & 96.37{\scriptsize$\pm$0.20} & 83.76{\scriptsize$\pm$0.08} \\
    \multicolumn{1}{c|}{} & & & 95.34{\scriptsize$\pm$0.30} & 82.85{\scriptsize$\pm$0.07} \\
    \midrule
    \multicolumn{3}{c|}{LoRA~\cite{lora}} & 95.96{\scriptsize$\pm$0.12} & 83.41{\scriptsize$\pm$0.11}  \\ 
    \bottomrule
  \end{tabular}
  }
  \label{tab:loss}
\end{table*}

\subsection{Ablation Study}

\begin{figure*}[!tbp]
  \centering
  \includegraphics[width=0.95\linewidth]{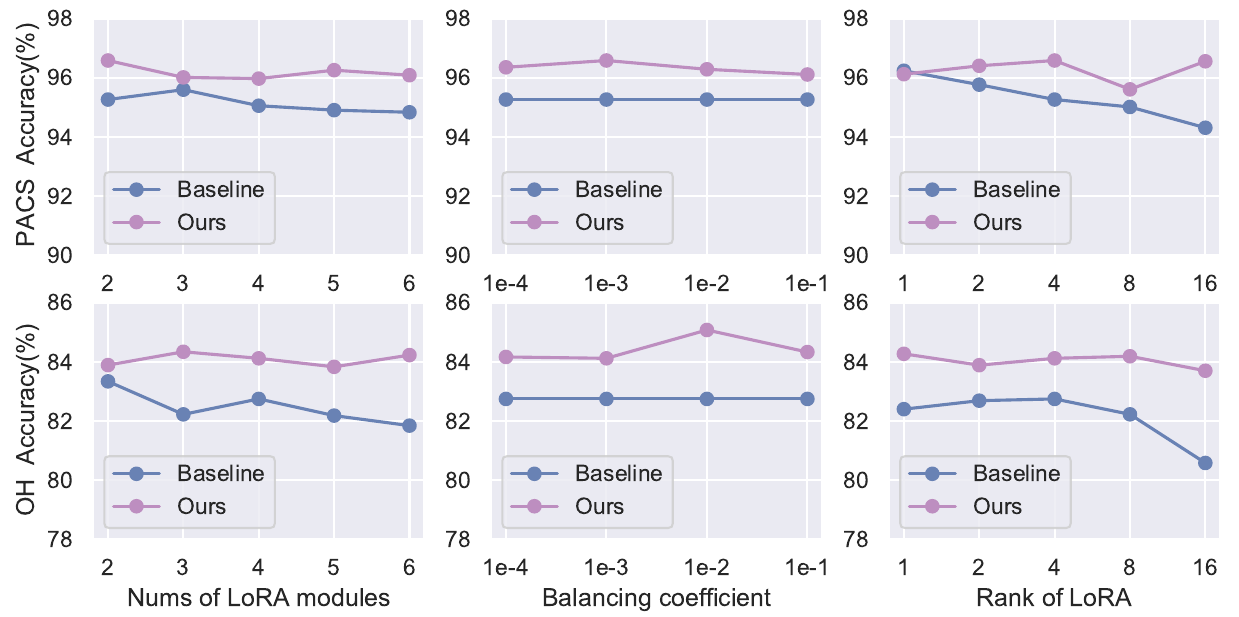}
  \caption{Leave-one-domain-out accuracy (\%) on PACS and OfficeHome when choosing different numbers of LoRA modules $N$, balancing coefficient $\alpha$ and rank of LoRA $r$. Baseline (blue line) indicates injecting a group of LoRA layers into the pre-trained model without applying $\mathcal{L}_{preserve}$ and $\mathcal{L}_{diversify}$ (\ie, balancing coefficient is zero).}
  \label{fig:ablation}
\end{figure*}

\textbf{Effectiveness of Two Orthogonal Losses.} 
To verify the improvement of the model's generalization performance by employing our proposed losses, we conduct ablation experiments about $\mathcal{L}_{preserve}$ and $\mathcal{L}_{diversify}$ on the PACS and OfficeHome benchmarks. As shown in \cref{tab:loss}, when $\mathcal{L}_{preserve}$ and $\mathcal{L}_{diversify}$ are applied simultaneously, the model achieves the best performance (blue row). We notice that the performance of injecting a group of LoRA layers into the pre-trained model without applying regularization loss ($4$th row) is worse than the performance of the original LoRA. This indicates that increasing the number of training parameters without regularization cannot improve the model's generalization performance, while our losses can effectively utilize more parameters.\\
\textbf{Effects of Numbers of LoRA Modules.}
Intuitively, the more LoRA modules in our method, the higher the probability of learning diverse knowledge. However, excessive modules lead to complicating loss optimization and increasing training overhead. The first column of \cref{fig:ablation} shows the performances of PEGO and Baseline when choosing different numbers of modules. PEGO achieves higher accuracy and is more stable than the Baseline.\\
\textbf{Effects of Balancing Coefficient.}
The second column of \cref{fig:ablation} shows the performances of Baseline and PEGO with different balancing coefficients. In a wide range of coefficients from 1e-4 to 1e-1, PEGO outperforms Baseline (balancing coefficient is zero), demonstrating the effectiveness of our proposed loss.\\
\textbf{Effects of Rank of LoRA.}
As shown in the third column of \cref{fig:ablation}, when the rank of the LoRA module is too high (greater than 8), the accuracy of Baseline significantly decreases, while the accuracy of our method remains stable. 

\begin{table*}[!tbp]
  \centering
  \caption{Performance comparison with Zero-shot CLIP. Leave-one-domain-out accuracy (\%) on four DomainBed benchmarks. The results of Zero-shot come from Lew \etal~\cite{GESTUR}. The best and second-best accuracy are \textbf{bolded} and \underline{underlined}, respectively.}
  \scalebox{0.93}{
  \begin{tabular}{p{2.3cm}<{\centering}|p{1.5cm}<{\centering}p{1.5cm}<{\centering}p{1.5cm}<{\centering}p{1.5cm}<{\centering}|p{0.8cm}<{\centering}}
    \toprule
    Algorithm & PACS & VLCS & OH & TI & Avg\\
    \midrule
    ERM (FT) & 83.4{\scriptsize$\pm$0.5} & 75.9{\scriptsize$\pm$1.3} & 66.4{\scriptsize$\pm$0.5} & \underline{35.3{\scriptsize$\pm$0.8}} & 65.3 \\
    Zero-shot~\cite{clip} & \textbf{96.8{\scriptsize$\pm$0.0}} & \underline{81.7{\scriptsize$\pm$0.3}} & \underline{83.0{\scriptsize$\pm$0.3}} & 31.3{\scriptsize$\pm$0.2} & \underline{73.2} \\
    \midrule
    \rowcolor{blue!10} Ours & \underline{96.5{\scriptsize$\pm$0.1}} & \textbf{83.2{\scriptsize$\pm$0.3}} & \textbf{84.2{\scriptsize$\pm$0.1}} & \textbf{57.3{\scriptsize$\pm$0.3}} & \textbf{80.3} \\
    \bottomrule
  \end{tabular}
  }
  \label{tab:zero_shot}
\end{table*}

\subsection{Comparison with the Zero-shot Baseline}
In \cref{sec:main_results}, we choose the CLIP~\cite{clip} pre-trained model as the default backbone. CLIP learns representations on 400 million image-text pairs and has demonstrated strong zero-shot ability on plenty of visual datasets.

\cref{tab:zero_shot} shows the performance comparison between our method and Zero-shot CLIP on four DomainBed benchmarks. In addition to being slightly lower than Zero-shot on the simple benchmark PACS by 0.3\%, PEGO achieves the best performances on the other three benchmarks. It outperforms zero-shot by 7.1\% in average performance (73.2\% $\rightarrow$ 80.3\%). Besides, we notice that the accuracy of Zero-shot on TerraIncognita is worse than ERM and our method surpasses Zero-shot by 26\% (31.3\% $\rightarrow$ 57.3\%). This result is consistent with the finding of Cho \etal~\cite{Promptstyler}. Although CLIP can leverage text information to achieve zero-shot without training data, we argue that source domain data is crucial for enhancing the generalization ability of the model and the key factor is whether robust fine-tuning can be accomplished.

\begin{table*}[!tbp]
  \centering
  \caption{Leave-one-domain-out accuracy (\%) on four DomainBed benchmarks when using ViT-L/14 pre-trained by CLIP as the backbone.}
  \scalebox{0.93}{
  \begin{tabular}{p{1.7cm}<{\centering}|p{1.5cm}<{\centering}p{1.5cm}<{\centering}p{1.5cm}<{\centering}p{1.5cm}<{\centering}|p{0.8cm}<{\centering}}
    \toprule
    Algorithm & PACS & VLCS & OH & TI & Avg\\
    \midrule
    ERM & 88.0{\scriptsize$\pm$4.1} & 77.5{\scriptsize$\pm$0.6} & 53.0{\scriptsize$\pm$3.2} & 43.3{\scriptsize$\pm$0.7} & 65.5 \\
    LoRA~\cite{lora} & \textbf{98.1{\scriptsize$\pm$0.0}} & \textbf{83.7{\scriptsize$\pm$0.3}} & 87.9{\scriptsize$\pm$0.1} & 52.7{\scriptsize$\pm$0.8} & 80.6 \\
    \midrule
    \rowcolor{blue!10} PEGO & 98.0{\scriptsize$\pm$0.1} & \textbf{83.7{\scriptsize$\pm$0.2}} & \textbf{88.6{\scriptsize$\pm$0.1}} & \textbf{57.2{\scriptsize$\pm$0.5}} & \textbf{81.9} \\
    \bottomrule
  \end{tabular}
  }
  \label{tab:backbone}
\end{table*}

\begin{figure*}[!tbp]
  \centering
  \includegraphics[width=0.77\linewidth]{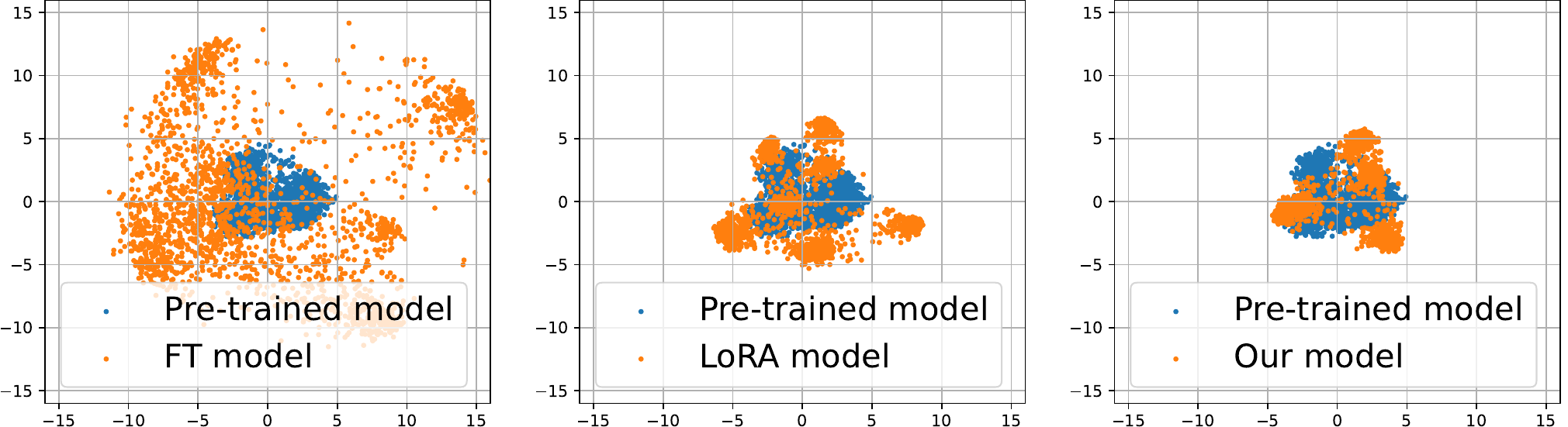}
  \caption{The visualization of the feature space (before the classifier) extracted by FT model, pre-trained model, LoRA model, and our model when training the PACS dataset and the test domain is \textit{art painting}.}
  \label{fig:features}
\end{figure*}

\subsection{Experiment Using ViT-L as the Backbone}
\label{sec:large}
In \cref{sec:main_results}, we choose ViT-B/16~\cite{vit} pre-trained by CLIP~\cite{clip} as the backbone for all the experiments. To verify the effectiveness of our method on larger models, we conduct the experiment using ViT-L/14 pre-trained by CLIP as the backbone. \cref{tab:backbone} provides the performances of ERM, LoRA~\cite{lora} and PEGO on four DomainBed benchmarks. Both LoRA and PEGO outperform ERM significantly on all the benchmarks and achieve similar accuracy on PACS and VLCS. However, on the other two benchmarks, PEGO has a significant improvement compared to LoRA, especially on TerraIncognita (52.7\% $\rightarrow$ 57.2\%).

\subsection{Visualization of Feature Space}
To understand whether our method can ``\textbf{\textit{Learn to Preserve}}'', we visualize the difference in feature space between our method and the pre-trained model with PCA~\cite{PCA} and compare with full fine-tuning (FT) and LoRA. As shown in \cref{fig:features}, fine-tuning all layers largely distorts the feature distribution and LoRA can partially alleviate it. 
Instead, our method successfully preserves pre-trained features by further using our orthogonal loss $\mathcal{L}_{preserve}$.

\begin{figure*}[!tbp]
  \centering
  \includegraphics[width=0.9\linewidth]{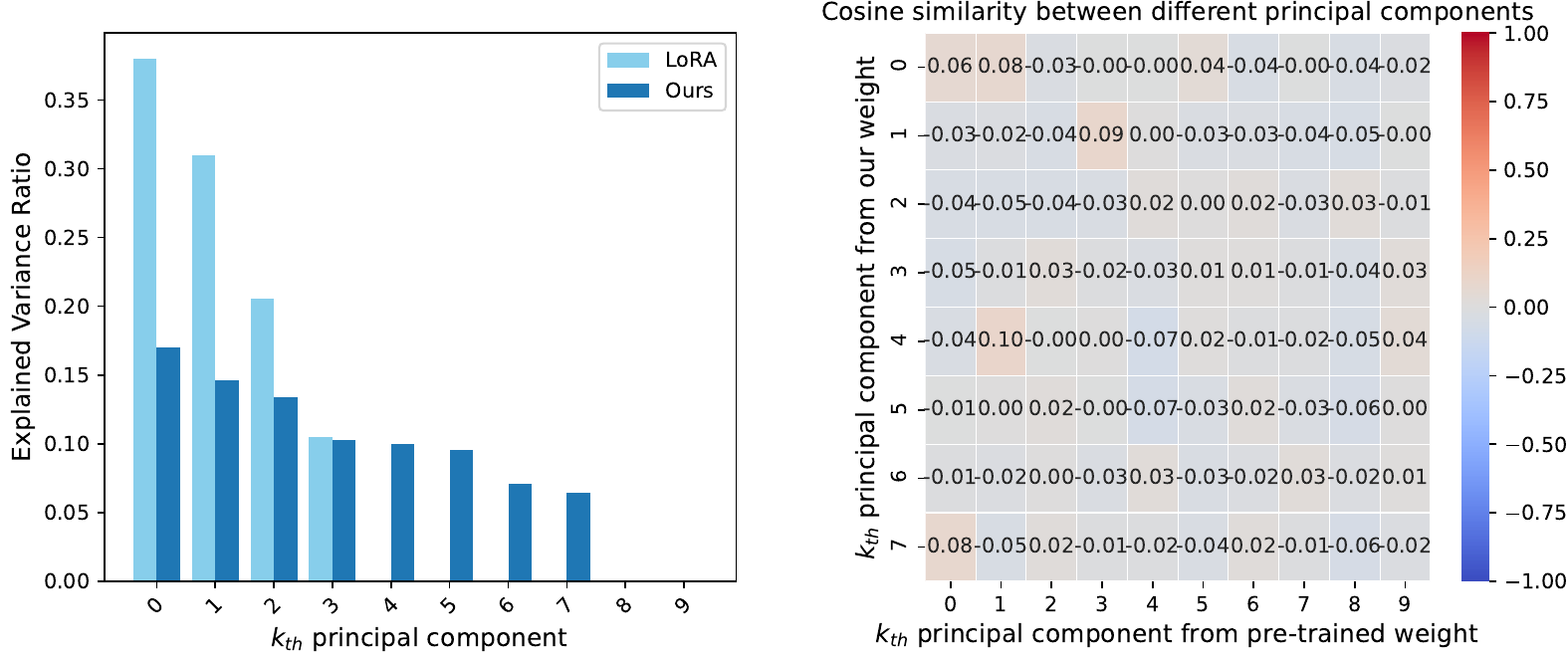}
  \caption{\textbf{Left:} Explained Variance Ratio of the top-10 PCs in LoRA weight and the top-10 PCs in PEGO weight. \textbf{Right:} Cosine similarity between the top-10 PCs of pre-trained weight and the top-8 PCs of PEGO weight.}
  \label{fig:PCA}
\end{figure*}

\subsection{Principal Component Analysis on Weights}
To confirm that our method achieves orthogonal regularization, we experiment by decomposing the model weights to get their principal components (PCs). Specifically, to verify the effect of  ``\textbf{\textit{Learn to Preserve}}'', we expect the learned PCs to be orthogonal to the PCs of pre-trained weights, and to verify ``\textbf{\textit{Learn to Diversify}}'', we expect the learned PCs to be more than the original LoRA. 
As shown in \cref{fig:PCA}, the weight of our model has more PCs (8 \vs 4) than LoRA and also exhibits orthogonal (zero cosine similarity) to the PCs of the pre-trained weights, validating the effectiveness of our proposed two losses.

\section{Conclusion}
In this paper, we address the problem of using foundation models in DG from a novel perspective of Learning to Preserve and Diversify. Specifically, we propose Parameter-Efficient Group with Orthogonal regularization (PEGO), which effectively preserves the generalization ability of pre-trained models and learns diverse knowledge. We conduct comparative experiments and ablation experiments to demonstrate the effectiveness and stability of PEGO. Our simple method can be applied to any neural network architecture with linear layers and is training-friendly without additional testing costs.


\section*{Acknowledgements}
The work is supported by the NSFC Project (62222604, 62206052, 62192783), Jiangsu Natural Science Foundation Project (BK20210224), China Postdoctoral Science Foundation (2024M750424), the Fundamental Research Funds for the Central Universities (020214380120), and the State Key Laboratory Funds for Key Project (ZZKT2024A14).

%
%
\bibliographystyle{splncs04}
\bibliography{main}

\appendix

\title{Learn to Preserve and Diversify: Parameter-Efficient Group with Orthogonal Regularization for Domain Generalization\\ ---Supplementary---} 

\titlerunning{Parameter-Efficient Group with Orthogonal Regularization for DG}

\author{Jiajun Hu\inst{1,2},
Jian Zhang\inst{1,2},
Lei Qi\inst{3,*},
Yinghuan Shi\inst{1,2,*},
Yang Gao\inst{1,2}}

\authorrunning{J.~Hu et al.}

\institute{State Key Laboratory for Novel Software Technology, Nanjing University, China \and
National Institute of Healthcare Data Science, Nanjing University, China \and
School of Computer Science and Engineering, Key Lab of Computer Network and Information Integration (Ministry of Education), Southeast University, China}

\maketitle
\renewcommand\thesection{\Alph{section}}

\section{Algorithm}
\begin{algorithm}[!htbp]
\caption{Parameter-Efficient Group with Orthogonal Regularization}
\label{alg:algorithm}
    \begin{algorithmic}[1]
        \Require training data $D_{tr}$, pre-trained vision transformer $F(\cdot;\theta)$ with $B$ blocks, classification head $H(\cdot;\psi)$, group of LoRA modules $g(\cdot;\phi)$, balancing coefficient $\alpha$, iteration $T$
        \State \textbf{Initialization:} Inject $g(\cdot;\phi)$ into $F(\cdot;\theta)$ to get the pre-trained model with group of LoRA modules $G(\cdot;\Phi)$ and freeze the pre-trained model weight
        \For{$t$ = 1, 2, ..., $T$}
            \State sample a batch $(x, y)$ in $D_{tr}$
            \State $\mathcal{L}_{cls} \leftarrow \mathcal{L}_{CE}(H(G(x; \Phi);\psi), y)$ \Comment{Eq. (2)}
            \State  $\mathcal{L}_{OR} \leftarrow 0$
            \For{$b$ = 1, 2, ..., $B$}
                \State $\mathcal{L}_{OR} \leftarrow \mathcal{L}_{OR}$ + $(\mathcal{L}_{O}(W^{q}_{b}) + \mathcal{L}_{O}(W^{v}_{b}))$ \Comment{Eq. (8)}
            \EndFor
            \State $\mathcal{L}_{final} \leftarrow \mathcal{L}_{cls} + \alpha\mathcal{L}_{OR}$ \Comment{Eq. (9)}
            \State update $g(\cdot;\phi)$, $H(\cdot;\psi)$ to minimize $\mathcal{L}$.
        \EndFor
        \State Merge the LoRA group with the pre-trained weight. \Comment{Eq. (10)}
        \State \Return $G$, $H$
    \end{algorithmic}
\end{algorithm}

\section{Evaluation Protocol and Hyperparameters Search}
In this section, we provide a detailed description of our evaluation protocol and hyperparameters (HPs) search. 
In line with prior research in DG, we designate one domain within the dataset as the unseen test domain, while the remaining domains serve as source domains. The final experimental results are obtained by averaging the accuracies across all test domains. To maintain consistency with DomainBed~\cite{domainbed}, $20\%$ of the samples from each source domain are allocated for validation and we adopt the training-domain validation strategy for hyperparameter search and model selection. Furthermore, all experiments are conducted using three different random seeds to ensure the reliability and reproducibility of our experiments.

As for algorithm-agnostic HPs in DomainBed (\eg, learning rate, dropout, weight decay), to reduce the training overhead caused by HPs search, we do not tune any algorithm-agnostic HPs. Specifically, for all the experiments, the learning rate, dropout, and weight decay are fixed to 5e-4, 0, and 0.
As regards the algorithm-specific HPs, we fix the rank of LoRA~\cite{lora} $r$ to 4 and the balance coefficient $\alpha$ to 1e-3 for all the experiments. We only search for the number of LoRA modules $N$ from $\{2, 4, 6\}$. \cref{tab:hp} provides a summary of the searched hyperparameter $N$ on five DomainBed benchmarks in our experiments.

\begin{table*}[!tbp]
  \centering
  \caption{The hyperparameter $N$ used on five DomainBed benchmarks in our experiments.}
  \scalebox{1.0}{
  \begin{tabular}{p{2.5cm}<{\centering}|p{1.2cm}<{\centering}p{1.2cm}<{\centering}p{1.2cm}<{\centering}p{1.2cm}<{\centering}p{1.2cm}<{\centering}}
    \toprule
    Hyperparameter & PACS & VLCS & OH & TI & DN \\
    \midrule
    $N$ & 2 & 4 & 4 & 4 & 4 \\
    \bottomrule
  \end{tabular}
  }
  \label{tab:hp}
\end{table*}

As shown in the ablation experiments of the main body (Sec. 5.1, Pages 12-13), the performance of our method is not sensitive to algorithm-specific HPs. 
Besides, to save GPU memory, we use half-precision (FP16) during training and inference for all the experiments. 


\begin{table*}[!tbp]
  \centering
  \caption{Performance comparison with more methods. Leave-one-domain-out accuracy (\%) on five DomainBed benchmarks.}
  \scalebox{1.0}{
  \begin{tabular}{p{2.3cm}<{\centering}|p{1.5cm}<{\centering}p{1.5cm}<{\centering}p{1.5cm}<{\centering}p{1.5cm}<{\centering}p{1.5cm}<{\centering}|p{0.8cm}<{\centering}}
    \toprule
    Algorithm & PACS & VLCS & OH & TI & DN & Avg\\
    \midrule
    Auto-RGN~\cite{auto} & 90.3{\scriptsize$\pm$0.5} & 80.7{\scriptsize$\pm$0.3} & 76.7{\scriptsize$\pm$0.5} & 48.5{\scriptsize$\pm$0.6} & 51.2{\scriptsize$\pm$0.7} & 69.5 \\
    CoOp~\cite{CoOp} & 96.1{\scriptsize$\pm$0.2} & 80.5{\scriptsize$\pm$0.6} & 84.2{\scriptsize$\pm$0.1} & 49.4{\scriptsize$\pm$0.6} & 59.3{\scriptsize$\pm$0.1} & 73.9 \\
    UPT~\cite{UPT} & \textbf{96.5{\scriptsize$\pm$0.2}} & 82.7{\scriptsize$\pm$0.1} & \textbf{84.4{\scriptsize$\pm$0.2}} & 54.9{\scriptsize$\pm$0.9} & \textbf{60.2{\scriptsize$\pm$0.1}} & 75.7 \\
    \midrule
    \rowcolor{blue!10} PEGO & \textbf{96.5{\scriptsize$\pm$0.1}} & \textbf{83.2{\scriptsize$\pm$0.3}} & 84.2{\scriptsize$\pm$0.1} & \textbf{57.3{\scriptsize$\pm$0.3}} & 59.3{\scriptsize$\pm$0.1} &
    \textbf{76.1} \\
    \bottomrule
  \end{tabular}
  }
  \label{tab:more_methods}
\end{table*}

\section{Comparisons with More Methods}
In this subsection, we conduct a performance comparison between PEGO and more methods, including Auto-RGN~\cite{auto}, CoOp~\cite{CoOp}, and UPT~\cite{UPT}. Auto-RGN measures the Relative Gradient Norm (RGN) of each transformer layer and sets different learning rates for each layer by its RGN. CoOp and UPT are both Prompt Learning methods that introduce learnable text or visual prompts for fine-tuning. As shown in \cref{tab:more_methods}, our method achieves better average performance than other methods benefiting from the proposed preserving and diversifying losses.

\begin{table*}[!tbp]
  \centering
  \caption{Trainable Parameters of Different Methods.}
  \scalebox{0.86}{
  \setlength{\tabcolsep}{0.5mm}{
      \begin{tabular}{c|c|c|c|c|c|c|c|c}
        \toprule
        & FT & Adapter\cite{adapter} & LoRA\cite{lora} & VPT\cite{vpt} & CoOp\cite{CoOp} & UPT\cite{UPT} & Auto-RGN\cite{auto} & PEGO\\
        \midrule
        Parameters & 86M & 0.16M & 0.15M & 0.10M & 2048 & 0.57M & 86M &0.29M \\
        \bottomrule
      \end{tabular}
      }
  }
  \label{tab:parameters}
\end{table*}

\section{Trainable Parameters of Different Methods }
The trainable parameters for each dataset are different due to the dimension difference of the classifier. We compare the trainable parameters of all methods on the PACS dataset. As shown in \cref{tab:parameters}, our method is significantly parameter-efficient compared to FT (0.29M \vs 86M).

\section{Detail Results of Each Domain}
\label{sec:each}
In this section, \cref{tab:PACS,tab:VLCS,tab:OH,tab:Terra,tab:dn} provide the detailed accuracy of algorithms on five DomainBed~\cite{domainbed} benchmarks: PACS~\cite{pacs}, VLCS~\cite{vlcs}, OfficeHome~\cite{officehome}, TerraIncognita~\cite{terra} and DomainNet~\cite{domainnet}. Since SWAD~\cite{swad}, SMA~\cite{SMA}, and GESTUR~\cite{GESTUR} do not report the detailed results of each domain in their papers, we only present the results of ERM, MIRO~\cite{miro}, Adapter~\cite{adapter}, LoRA~\cite{lora}, VPT~\cite{vpt}, L$^{2}$-SP~\cite{L2SP}, LP-FT~\cite{LPFT}, LwF~\cite{LwF} and PEGO.\\ \\

\begin{table*}[!tbp]
  \centering
  \caption{Leave-one-domain-out accuracy (\%) of each domain on PACS when using ViT-B/16 pre-trained by CLIP as the backbone.}
  \scalebox{1.0}{
  \begin{tabular}{p{1.8cm}<{\centering}|p{1.5cm}<{\centering}p{1.5cm}<{\centering}p{1.5cm}<{\centering}p{1.5cm}<{\centering}|p{1.5cm}<{\centering}}
    \toprule
    Algorithm & A & C & P & S & Avg\\
    \midrule
    ERM (FT) & 80.5{\scriptsize$\pm$3.4} & 86.4{\scriptsize$\pm$0.6} & 93.4{\scriptsize$\pm$1.0} & 73.2{\scriptsize$\pm$3.9} & 83.4{\scriptsize$\pm$0.4} \\
    \midrule
    MIRO~\cite{miro} & 95.6{\scriptsize$\pm$0.6} & 96.6{\scriptsize$\pm$0.2} & 99.7{\scriptsize$\pm$0.1} & 90.7{\scriptsize$\pm$2.5} & 95.6{\scriptsize$\pm$0.6} \\
    \midrule
    Adapter~\cite{adapter} & 91.8{\scriptsize$\pm$0.2} & 93.1{\scriptsize$\pm$0.4} & 98.8{\scriptsize$\pm$0.1} & 84.4{\scriptsize$\pm$1.6} & 92.0{\scriptsize$\pm$0.5} \\
    LoRA~\cite{lora} & \textbf{97.4{\scriptsize$\pm$0.3}} & 97.5{\scriptsize$\pm$0.1} & 99.7{\scriptsize$\pm$0.1} & 89.2{\scriptsize$\pm$0.4} & 96.0{\scriptsize$\pm$0.1} \\
    VPT~\cite{vpt} & 97.1{\scriptsize$\pm$0.4} & 97.8{\scriptsize$\pm$0.1} & \textbf{99.9{\scriptsize$\pm$0.0}} & 90.1{\scriptsize$\pm$0.9} & 96.2{\scriptsize$\pm$0.3} \\
    \midrule
    L$^{2}$-SP~\cite{L2SP} & 93.9{\scriptsize$\pm$1.0} & 94.3{\scriptsize$\pm$0.6} & 97.8{\scriptsize$\pm$0.3} & 83.1{\scriptsize$\pm$2.3} & 92.2{\scriptsize$\pm$0.7} \\
    LwF~\cite{LwF} & 93.2{\scriptsize$\pm$1.4} & 94.2{\scriptsize$\pm$0.7} & 98.5{\scriptsize$\pm$0.2} & 88.8{\scriptsize$\pm$0.4} & 93.6{\scriptsize$\pm$0.6} \\
    LP-FT~\cite{LPFT} & 89.1{\scriptsize$\pm$2.8} & 97.8{\scriptsize$\pm$0.1} & 99.8{\scriptsize$\pm$0.0} & 89.9{\scriptsize$\pm$0.2} & 94.2{\scriptsize$\pm$0.7} \\
    \midrule
    \rowcolor{blue!10} PEGO & 97.1{\scriptsize$\pm$0.1} & \textbf{98.5{\scriptsize$\pm$0.2}} & 99.7{\scriptsize$\pm$0.1} & \textbf{90.9{\scriptsize$\pm$0.2}} & \textbf{96.5{\scriptsize$\pm$0.1}} \\
    \bottomrule
  \end{tabular}
  }
  \label{tab:PACS}
\end{table*}

\begin{table*}[!tbp]
  \centering
  \caption{Leave-one-domain-out accuracy (\%) of each domain on VLCS when using ViT-B/16 pre-trained by CLIP as the backbone.}
  \scalebox{1.0}{
  \begin{tabular}{p{1.8cm}<{\centering}|p{1.5cm}<{\centering}p{1.5cm}<{\centering}p{1.5cm}<{\centering}p{1.5cm}<{\centering}|p{1.5cm}<{\centering}}
    \toprule
    Algorithm & C & L & S & V & Avg\\
    \midrule
    ERM (FT) & 95.4{\scriptsize$\pm$0.6} & 65.6{\scriptsize$\pm$0.9} & 72.9{\scriptsize$\pm$2.2} & 69.9{\scriptsize$\pm$2.2} & 75.9{\scriptsize$\pm$1.1} \\
    \midrule
    MIRO~\cite{miro} & 98.9{\scriptsize$\pm$0.5} & 67.1{\scriptsize$\pm$1.0} & 81.9{\scriptsize$\pm$0.4} & 81.2{\scriptsize$\pm$0.2} & 82.3{\scriptsize$\pm$0.2} \\
    \midrule
    Adapter~\cite{adapter} & 95.7{\scriptsize$\pm$0.2} & 65.9{\scriptsize$\pm$0.9} & 79.5{\scriptsize$\pm$0.7} & 78.0{\scriptsize$\pm$0.7} & 79.8{\scriptsize$\pm$0.4} \\
    LoRA~\cite{lora} & 96.1{\scriptsize$\pm$0.4} & \textbf{68.1{\scriptsize$\pm$0.2}} & 83.5{\scriptsize$\pm$0.3} & 83.1{\scriptsize$\pm$0.4} & 82.7{\scriptsize$\pm$0.0} \\
    VPT~\cite{vpt} & 96.8{\scriptsize$\pm$0.5} & 67.2{\scriptsize$\pm$0.3} & \textbf{84.9{\scriptsize$\pm$0.2}} & 82.6{\scriptsize$\pm$0.4} & 82.9{\scriptsize$\pm$0.3} \\
    \midrule
    LP-FT~\cite{LPFT} & 94.5{\scriptsize$\pm$0.3} & 62.0{\scriptsize$\pm$0.3} & 76.4{\scriptsize$\pm$1.3} & 77.0{\scriptsize$\pm$2.9} & 77.5{\scriptsize$\pm$0.4} \\
    L$^{2}$-SP~\cite{L2SP} & 96.8{\scriptsize$\pm$0.9} & 66.2{\scriptsize$\pm$1.0} & 78.5{\scriptsize$\pm$1.6} & 82.5{\scriptsize$\pm$0.2} & 81.0{\scriptsize$\pm$0.2} \\
    LwF~\cite{LwF} & \textbf{99.1{\scriptsize$\pm$0.3}} & 65.5{\scriptsize$\pm$1.4} & 80.4{\scriptsize$\pm$1.2} & 82.6{\scriptsize$\pm$0.2} & 81.9{\scriptsize$\pm$0.4} \\
    \midrule
    \rowcolor{blue!10} PEGO & 96.4{\scriptsize$\pm$0.1} & 67.8{\scriptsize$\pm$0.5} & 83.3{\scriptsize$\pm$0.3} & \textbf{85.2{\scriptsize$\pm$1.0}} & \textbf{83.2{\scriptsize$\pm$0.3}} \\
    \bottomrule
  \end{tabular}
  }
  \label{tab:VLCS}
\end{table*}

\begin{table*}[!tbp]
  \centering
  \caption{Leave-one-domain-out accuracy (\%) of each domain on OfficeHome when using ViT-B/16 pre-trained by CLIP as the backbone.}
  \scalebox{1.0}{
  \begin{tabular}{p{1.8cm}<{\centering}|p{1.5cm}<{\centering}p{1.5cm}<{\centering}p{1.5cm}<{\centering}p{1.5cm}<{\centering}|p{1.5cm}<{\centering}}
    \toprule
    Algorithm & A & C & P & R & Avg\\
    \midrule
    ERM (FT) & 59.2{\scriptsize$\pm$1.3} & 56.1{\scriptsize$\pm$0.6} & 74.8{\scriptsize$\pm$0.1} & 75.4{\scriptsize$\pm$0.8} & 66.4{\scriptsize$\pm$0.4} \\
    \midrule
    MIRO~\cite{miro} & 80.8{\scriptsize$\pm$0.1} & 72.2{\scriptsize$\pm$0.5} & 88.6{\scriptsize$\pm$0.3} & 88.5{\scriptsize$\pm$0.2} & 82.5{\scriptsize$\pm$0.1} \\
    \midrule
    Adapter~\cite{adapter} & 67.1{\scriptsize$\pm$1.2} & 61.7{\scriptsize$\pm$0.4} & 81.5{\scriptsize$\pm$0.5} & 81.3{\scriptsize$\pm$0.6} & 72.9{\scriptsize$\pm$0.4} \\
    LoRA~\cite{lora} & 83.2{\scriptsize$\pm$0.2} & 71.8{\scriptsize$\pm$0.4} & 89.1{\scriptsize$\pm$0.2} & \textbf{89.5{\scriptsize$\pm$0.2}} & 83.4{\scriptsize$\pm$0.1} \\
    VPT~\cite{vpt} & 82.9{\scriptsize$\pm$0.6} & 71.5{\scriptsize$\pm$0.6} & 89.7{\scriptsize$\pm$0.1} & \textbf{89.5{\scriptsize$\pm$0.3}} & 83.4{\scriptsize$\pm$0.3} \\
    \midrule
    L$^{2}$-SP~\cite{L2SP} & 62.6{\scriptsize$\pm$1.3} & 57.1{\scriptsize$\pm$0.4} & 76.4{\scriptsize$\pm$0.8} & 76.6{\scriptsize$\pm$0.2} & 68.2{\scriptsize$\pm$0.5} \\
    LP-FT~\cite{LPFT} & 64.5{\scriptsize$\pm$1.4} & 68.0{\scriptsize$\pm$0.4} & 76.7{\scriptsize$\pm$0.3} & 79.0{\scriptsize$\pm$0.2} & 72.0{\scriptsize$\pm$0.4} \\
    LwF~\cite{LwF} & 79.0{\scriptsize$\pm$1.7} & 70.4{\scriptsize$\pm$0.7} & 86.8{\scriptsize$\pm$0.3} & 86.7{\scriptsize$\pm$0.4} & 80.7{\scriptsize$\pm$0.4} \\
    \midrule
    \rowcolor{blue!10} PEGO & \textbf{83.7{\scriptsize$\pm$0.3}} & \textbf{73.3{\scriptsize$\pm$0.4}} & \textbf{90.3{\scriptsize$\pm$0.3}} & \textbf{89.5{\scriptsize$\pm$0.3}} & \textbf{84.2{\scriptsize$\pm$0.1}} \\
    \bottomrule
  \end{tabular}
  }
  \label{tab:OH}
\end{table*}

\begin{table*}[!tbp]
  \centering
  \caption{Leave-one-domain-out accuracy (\%) of each domain on TerraIncognita when using ViT-B/16 pre-trained by CLIP as the backbone.}
  \scalebox{1.0}{
  \begin{tabular}{p{1.8cm}<{\centering}|p{1.5cm}<{\centering}p{1.5cm}<{\centering}p{1.5cm}<{\centering}p{1.5cm}<{\centering}|p{1.5cm}<{\centering}}
    \toprule
    Algorithm & L100 & L38 & L43 & L46 & Avg\\
    \midrule
    ERM (FT) & 38.1{\scriptsize$\pm$0.3} & 26.7{\scriptsize$\pm$2.5} & 41.9{\scriptsize$\pm$1.3} & 34.4{\scriptsize$\pm$1.8} & 35.3{\scriptsize$\pm$0.6} \\
    \midrule
    MIRO~\cite{miro} & \textbf{65.0{\scriptsize$\pm$0.6}} & 46.7{\scriptsize$\pm$0.7} & 60.8{\scriptsize$\pm$1.3} & 44.9{\scriptsize$\pm$0.1} & 54.3{\scriptsize$\pm$0.3} \\
    \midrule
    Adapter~\cite{adapter} & 38.8{\scriptsize$\pm$5.1} & 44.9{\scriptsize$\pm$2.0} & 56.2{\scriptsize$\pm$0.3} & 37.8{\scriptsize$\pm$1.3} & 44.4{\scriptsize$\pm$0.8} \\
    VPT~\cite{vpt} & 55.0{\scriptsize$\pm$3.9} & 52.6{\scriptsize$\pm$1.3} & 61.3{\scriptsize$\pm$0.4} & 47.8{\scriptsize$\pm$0.4} & 54.2{\scriptsize$\pm$0.7} \\
    LoRA~\cite{lora} & 54.6{\scriptsize$\pm$2.4} & 52.7{\scriptsize$\pm$1.2} & 61.2{\scriptsize$\pm$0.8} & \textbf{50.5{\scriptsize$\pm$0.5}} & 54.8{\scriptsize$\pm$0.6} \\
    \midrule
    LP-FT~\cite{LPFT} & 42.8{\scriptsize$\pm$4.2} & 33.2{\scriptsize$\pm$3.3} & 46.7{\scriptsize$\pm$1.1} & 33.2{\scriptsize$\pm$1.1} & 39.0{\scriptsize$\pm$1.5} \\
    L$^{2}$-SP~\cite{L2SP} & 45.6{\scriptsize$\pm$5.5} & 27.2{\scriptsize$\pm$3.5} & 49.9{\scriptsize$\pm$1.3} & 34.8{\scriptsize$\pm$0.3} & 39.4{\scriptsize$\pm$1.6} \\
    LwF~\cite{LwF} & 44.4{\scriptsize$\pm$1.8} & 34.9{\scriptsize$\pm$2.6} & 47.5{\scriptsize$\pm$1.3} & 30.9{\scriptsize$\pm$3.8} & 39.4{\scriptsize$\pm$0.6} \\
    \midrule
    \rowcolor{blue!10} PEGO & 63.2{\scriptsize$\pm$0.3} & \textbf{56.4{\scriptsize$\pm$0.3}} & \textbf{61.8{\scriptsize$\pm$1.0}} & 47.9{\scriptsize$\pm$0.5} & \textbf{57.3{\scriptsize$\pm$0.3}} \\
    \bottomrule
  \end{tabular}
  }
  \label{tab:Terra}
\end{table*}

\begin{table*}[t]
  \centering
  \caption{Leave-one-domain-out accuracy (\%) of each domain on DomainNet when using ViT-B/16 pre-trained by CLIP as the backbone.}
  \scalebox{1.0}{
  \begin{tabular}{p{1.8cm}<{\centering}|p{1.3cm}<{\centering}p{1.3cm}<{\centering}p{1.3cm}<{\centering}p{1.3cm}<{\centering}p{1.3cm}<{\centering}p{1.3cm}<{\centering}|p{1.3cm}<{\centering}}
    \toprule
    Algorithm & clipart & infograph & painting & quickdraw & real & sketch & Avg\\
    \midrule
    ERM (FT) & 68.0{\scriptsize$\pm$0.1} & 22.5{\scriptsize$\pm$0.4} & 46.5{\scriptsize$\pm$2.4} & 18.5{\scriptsize$\pm$0.6} & 58.7{\scriptsize$\pm$1.6} & 52.5{\scriptsize$\pm$0.7} & 44.4{\scriptsize$\pm$0.5} \\
    \midrule
    MIRO~\cite{miro} & 74.9{\scriptsize$\pm$0.1} & 37.1{\scriptsize$\pm$0.2} & 59.8{\scriptsize$\pm$0.4} & 18.7{\scriptsize$\pm$0.8} & 72.2{\scriptsize$\pm$0.1} & 61.2{\scriptsize$\pm$0.6} & 54.0{\scriptsize$\pm$0.2} \\
    \midrule
    Adapter~\cite{adapter} & 75.6{\scriptsize$\pm$0.2} & 37.6{\scriptsize$\pm$0.2} & 63.1{\scriptsize$\pm$0.2} & 19.4{\scriptsize$\pm$0.3} & 77.2{\scriptsize$\pm$0.1} & 64.2{\scriptsize$\pm$0.3} & 56.2{\scriptsize$\pm$0.1} \\
    LoRA~\cite{lora} & 76.4{\scriptsize$\pm$0.1} & 43.3{\scriptsize$\pm$0.3} & 63.6{\scriptsize$\pm$0.3} & \textbf{19.5{\scriptsize$\pm$0.3}} & 79.2{\scriptsize$\pm$0.1} & 66.4{\scriptsize$\pm$0.1} & 58.1{\scriptsize$\pm$0.1} \\
    VPT~\cite{vpt} & 76.7{\scriptsize$\pm$0.0} & 43.1{\scriptsize$\pm$0.3} & 66.6{\scriptsize$\pm$0.1} & 19.4{\scriptsize$\pm$0.2} & 80.3{\scriptsize$\pm$0.0} & 67.4{\scriptsize$\pm$0.1} & 58.9{\scriptsize$\pm$0.1} \\
    \midrule
    LP-FT~\cite{LPFT} & 70.9{\scriptsize$\pm$0.2} & 26.7{\scriptsize$\pm$0.3} & 55.8{\scriptsize$\pm$0.3} & 17.1{\scriptsize$\pm$0.5} & 66.3{\scriptsize$\pm$0.4} & 57.5{\scriptsize$\pm$0.4} & 49.1{\scriptsize$\pm$0.3} \\
    L$^{2}$-SP~\cite{L2SP} & 70.6{\scriptsize$\pm$0.1} & 28.4{\scriptsize$\pm$0.3} & 55.6{\scriptsize$\pm$0.5} & 18.3{\scriptsize$\pm$0.5} & 68.5{\scriptsize$\pm$0.4} & 58.4{\scriptsize$\pm$0.1} & 50.0{\scriptsize$\pm$0.2} \\
    LwF~\cite{LwF} & 73.2{\scriptsize$\pm$0.1} & 30.6{\scriptsize$\pm$0.3} & 58.0{\scriptsize$\pm$0.5} & 18.6{\scriptsize$\pm$0.4} & 69.1{\scriptsize$\pm$0.2} & 60.8{\scriptsize$\pm$0.0} & 51.7{\scriptsize$\pm$0.1} \\
    \midrule
    \rowcolor{blue!10} PEGO & \textbf{76.8{\scriptsize$\pm$0.1}} & \textbf{44.6{\scriptsize$\pm$0.2}} & \textbf{67.1{\scriptsize$\pm$0.3}} & 18.8{\scriptsize$\pm$0.2} & \textbf{80.5{\scriptsize$\pm$0.1}} & \textbf{67.7{\scriptsize$\pm$0.1}} & \textbf{59.3{\scriptsize$\pm$0.1}} \\
    \bottomrule
  \end{tabular}
  }
  \label{tab:dn}
\end{table*}

\section{Limitation}
\label{sec:Limitation}
Although our method cannot be easily applied to some traditional convolutional neural networks not containing linear layers (\eg, ResNet~\cite{resnet}), it can be applied to any type of Transformer~\cite{transformer} architecture, similar to LoRA. With the increasing number of Transformer-based architectures being proposed (\eg, ViT~\cite{vit}, ConViT~\cite{convit}, DeiT~\cite{DeiT}), our method exhibits a wide range of applications for these networks.

\end{document}